\documentclass[sigconf]{acmart}

\usepackage{graphicx}
\usepackage{microtype}
\usepackage{hyperref}
\usepackage[capitalise]{cleveref}
\usepackage{booktabs}
\usepackage{tabularx}
\usepackage[british]{babel}
\usepackage{csquotes}
\usepackage[nolist]{acronym}
\usepackage[separate-uncertainty=true]{siunitx}

\copyrightyear{2022} 
\acmYear{2022} 
\setcopyright{acmlicensed}\acmConference[ICCAD '22]{IEEE/ACM International Conference on Computer-Aided Design}{October 30-November 3, 2022}{San Diego, CA, USA}
\acmBooktitle{IEEE/ACM International Conference on Computer-Aided Design (ICCAD '22), October 30-November 3, 2022, San Diego, CA, USA}
\acmPrice{15.00}
\acmDOI{10.1145/3508352.3549329}
\acmISBN{978-1-4503-9217-4/22/10}

\newcolumntype{Y}{>{\centering\arraybackslash}X}
\newcolumntype{R}{>{\raggedleft\arraybackslash}X}

\begin{acronym}[SRAM]
	\acro{nn}[NN]{neural network}
	\acro{agn}[AGN]{additive Gaussian noise}
	\acro{sram}[SRAM]{Static Random Access Memory}
	\acro{mc}[MC]{Monte Carlo}
	\acro{sgd}[SGD]{Stochastic Gradient Descent}
	\acro{ilp}[ILP]{Integer Linear Programming}
	\acro{qat}[QAT]{Quantization-aware Training}
	\acro{ste}[STE]{Straight-Through Estimator}
	\acro{mre}[MRE]{Mean Relative Error}
	\acro{am}[AM]{approximate multiplier}
	\acro{cnn}[CNN]{convolutional neural network}
	\acro{fc}[FC]{Fully-Connected}
\end{acronym}

\begin{document}

\title{Combining Gradients and Probabilities for Heterogeneous Approximation of Neural Networks}

\author{Elias Trommer}
\email{elias.trommer@infineon.com}
\affiliation{%
	\institution{Infineon Technologies}
	\streetaddress{Königsbrücker Straße 180}
	\postcode{01099}
	\city{Dresden}
	\country{Germany}
}
\author{Bernd Waschneck}
\email{bernd.waschneck@infineon.com}
\affiliation{%
	\institution{Infineon Technologies}
	\streetaddress{Königsbrücker Straße 180}
	\postcode{01099}
	\city{Dresden}
	\country{Germany}
}
\author{Akash Kumar}
\email{akash.kumar@tu-dresden.de}
\affiliation{%
	\institution{Center for Advancing Electronics Dresden (cfaed)}
	\city{Dresden}
	\country{Germany}
}

\begin{abstract}
This work explores the search for heterogeneous \acl{am} configurations for \aclp{nn} that produce high accuracy and low energy consumption. We discuss the validity of \acl{agn} added to accurate \acl{nn} computations as a surrogate model for behavioral simulation of \aclp{am}. The continuous and differentiable properties of the solution space spanned by the \acl{agn} model are used as a heuristic that generates meaningful estimates of layer robustness without the need for combinatorial optimization techniques. Instead, the amount of noise injected into the accurate computations is learned during network training using backpropagation. A probabilistic model of the multiplier error is presented to bridge the gap between the domains; the model estimates the standard deviation of the \acl{am} error, connecting solutions in the \acl{agn} space to actual hardware instances. Our experiments show that the combination of heterogeneous approximation and \acl{nn} retraining reduces the energy consumption for multiplications by 70\% to 79\% for different ResNet variants on the CIFAR-10 dataset with a Top-1 accuracy loss below one percentage point. For the more complex Tiny ImageNet task, our VGG16 model achieves a \SI{53}{\percent} reduction in energy consumption with a drop in Top-5 accuracy of 0.5 percentage points. We further demonstrate that our error model can predict the parameters of an \acl{am} in the context of the commonly used \ac{agn} model with high accuracy. Our software implementation is available under \url{https://github.com/etrommer/agn-approx}.
\end{abstract}


\maketitle
\acresetall 

\section{Introduction}
The power consumption of \acp{nn} has long been a major obstacle for their deployment inside power-constrained edge devices~\cite{xu18}. Particularly their computational complexity is a concern. Approximate arithmetic units have been proposed by the research community to address these issues~\cite{lotric12, mrazek16, zhang15}: relaxing the constraints imposed on the accuracy of operations enables new optimizations of arithmetic hardware, allowing for improved latency, energy consumption and area usage. Because of the dominant impact of multiplications, most efforts have been focused on improving the performance of multipliers using approximation---a rationale that this work builds upon as well.\par
Recent findings in the field of quantization demonstrate that optimizing quantization bit widths individually for each layer can provide higher accuracy compared to solutions that uniformly quantize the entire network. While some of these approaches rely on reinforcement learning~\cite{wang19, elthakeb20}, an alternative route is the optimization of quantization parameters using gradient-based methods during network training~\cite{yamamoto21, li20, zhang18}. To the best of our knowledge, the only work that investigates a non-uniform (i.e. heterogeneous) approach to approximation of \acp{nn} is~\citet{mrazek19}, which uses a multi-objective genetic algorithm. By relying on behavioral simulation to evaluate a large number of candidate solutions, this method can not perform retraining, as it would render the already lengthy search procedure computationally intractable. Recent work by~\citet{delaparra20} demonstrates that solutions employing a single \ac{am} throughout the entire network can outperform heterogeneous solutions if lost accuracy is recovered using retraining. Naturally, this raises the question whether performance could be improved even further if heterogeneous approximation and retraining were combined. It is obvious that this requires a different approach to finding a heterogeneous multiplier configuration in order for the search procedure to be feasible. To address this problem, we propose a gradient-based search algorithm that is capable of finding high-quality heterogeneous \ac{am} configurations. The search results in only a single candidate solution (or a small set acquired by varying a hyperparameter) that can easily be retrained. The key novel contributions of this work are:
\begin{itemize}
\item An efficient method for jointly determining the robustness to approximate multiplications for each \ac{nn} layer during training. Through transformation into a more favorable solution space, our algorithm allows for a fast traversal of the heterogeneous multiplier assignment problem in a \ac{nn}.
\item A probabilistic error model that can give a precise estimate of the performance of an \ac{am} in a \ac{nn} layer in terms of recoverable and non-recoverable error. Besides knowledge of the difference between the accurate and approximate multiplication results for all operand combinations, no behavioral simulation is required. The model is data-driven and does not assume any particular distribution of input operands. This makes its performance agnostic to methods that impact these distributions such as pruning~\cite{han15, trommer21}, quantization~\cite{zhao19}, etc.
\end{itemize}
Our evaluation of several \ac{cnn} models on the CIFAR-10 and Tiny ImageNet datasets shows that our method consistently manages to push the boundary of energy efficiency and network performance for various networks. We also improve upon existing models that express \ac{am} properties as \ac{agn} parameters. This boosts the accuracy of \ac{agn} as a faster and simpler replacement for behavioral simulation of \acp{am}.
\section{Background and Related Work}
Due to the significant complexity of accurately simulating \acp{am} during the training procedure of \acp{nn}~\cite{vaverka20, delaparra20proxsim}, several works propose the use of random noise as a replacement for the inaccurate computations.~\citet{hammad19} use a model based on the multiplier's \ac{mre} to generate random data which perturbs the output of an accurate computation. To enable retraining without the need for hardware simulation, \citet{delaparra21} propose a data-driven noise model with higher granularity. The model constructs a noise tensor individually for each neuron by observing the approximation error on sample data. Similar to behavioral simulation, this model is not generalizable across \ac{am} instances because it only captures the dynamics of the \ac{am} it was constructed for.\par
Noisy intermediate results have also been used to demonstrate that the robustness in a \ac{nn} is not uniform but varies for different layers~\cite{cheney17}. These findings are corroborated by \citet{hanif18}, who determine the individual robustness of layers to approximation by injecting \ac{agn} into individual layers and observing the change in accuracy. By optimizing one layer at a time, the work does not take into account interdependencies between the robustness of layers. It also lacks a method that connects the robustness of a given layer to a concrete hardware instance. A similar method in the context of Capsule Networks is discussed by~\citet{marchisio20}. The proposed model for connecting a layer's robustness to noise with \ac{am} instances requires \ac{mc} simulations to be carried out for each combination of layer and \ac{am}. Other models that describe the error incurred by a single approximate multiplication were put forward by~\citet{mazahir17} and~\citet{ullah21}. These models, however, do not consider the compounding effects of multiple operations in a neural network.\par
To leverage the varying robustness of layers~\citet{mrazek19} demonstrate the use of a multi-objective evolutionary algorithm as a means to tackle the large search space of heterogeneous multiplier assignment in a \ac{nn}. This solution relies on the evaluation of numerous candidate solutions, requiring the use of a weight remapping scheme rather than retraining to recover the degraded accuracy. Even without retraining, the vast number of simulations makes the method prohibitively slow for non-trivial networks. The same is true for more recent approaches that optimize the network architecture itself for use with \acp{am}~\cite{pinos21}.


\section{Proposed Methodology}
We provide an analysis of the aggregate error at a neuron's output and conclude that \ac{agn} is a meaningful surrogate model for behavioral simulation of \acp{am}. Using the properties of this model, we demonstrate how to simultaneously optimize the amount of \ac{agn} across the entire network, considering the complex interactions between perturbations in different layers. An error model is developed in order to make the abstract \ac{agn} parameter comparable with the computation errors exhibited by concrete \ac{am} instances. Using the learned robustness parameters and the error model, we can determine which \acp{am} will produce the required accuracy and match appropriate \ac{am} instances based on each layer's individual sensitivity.
\subsection{Modeling approximate multiplication as noise}
\label{sec:noise}
The error imposed by a single approximate multiplication can be considered additive to the output of an accurate multiplication
\begin{equation}\label{eq:approx_prod}
\tilde{f}(x,w) = x \cdot w + e(x,w)
\end{equation}
where $e(x,w)$ is an error function that is unique to each \ac{am} instance. For a~\ac{nn} application we are, however, not concerned with the error of each individual operation, but with the \emph{aggregate} error over several multiplications. With the definition of the pre-activation output of an accurate neuron
\begin{equation}
y = \sum\limits_{i=1}^{n}x_i w_i + b \label{eq:neuron}
\end{equation}
\Cref{eq:approx_prod} can be substituted to obtain the output of the same neuron using approximate multiplication as:
\begin{align}
\tilde{y} 	&= \sum\limits_{i=1}^{n}x_i w_i + e(x_i, w_i) + b \\
			&= \underbrace{\sum\limits_{i=1}^{n}x_i w_i + b}_{\text{Accurate Neuron}} + \underbrace{\sum\limits_{i=1}^{n} e(x_i, w_i)}_{\text{Aggregate Error}}
\end{align}
Assuming that $x$ and $w$ exhibit sufficiently random properties, we conjecture that, as $n$ grows, the distribution of the aggregate error will converge to a normal distribution, thus:
\begin{align}
\tilde{y} &\approx \sum\limits_{i=1}^{n}x_i w_i + b + \mathcal{N}(\mu_e, \sigma_e)\\
&= \sum\limits_{i=1}^{n}x_i w_i + b + \mu_e + \sqrt{\sigma_e} \cdot \mathcal{N}(0,1)\label{eq:noise_model}
\end{align}
Furthermore, the systematic portion of the error $\mu_e$ will be absorbed by the bias or subsequent batch normalization for any non-degenerate case when the network is retrained to match the approximate configuration~\cite{tasoulas20} s.t. $b' = b - \mu_e$. To simulate the effect of approximation on a fully retrained network, we can therefore assume that $\mu_e = 0$.\par
Using this surrogate model instead of a behavioral simulation of the approximation error has two important benefits for the search procedure:
First, \ac{agn} can easily be constructed using primitives that are available in most common deep learning toolkits and does not require low-level integration of handwritten kernels. Instead, it leverages the highly optimized accurate kernel implementations maintained by the framework and therefore adds very little runtime overhead to the training procedure.
Secondly, instead of selecting instances with unique characteristics out of a discrete (and potentially large) set of \acp{am}, the \ac{agn} model captures their most relevant property---the non-recoverable error---in a single, continuous parameter. Only after solving the optimization problem in the more favorable \ac{agn} space, the found error robustness is used to select matching hardware instances. 
\subsection{Gradient-based Robustness Optimization}
\begin{figure}[tb]
	\centering
	\includegraphics[width=\linewidth]{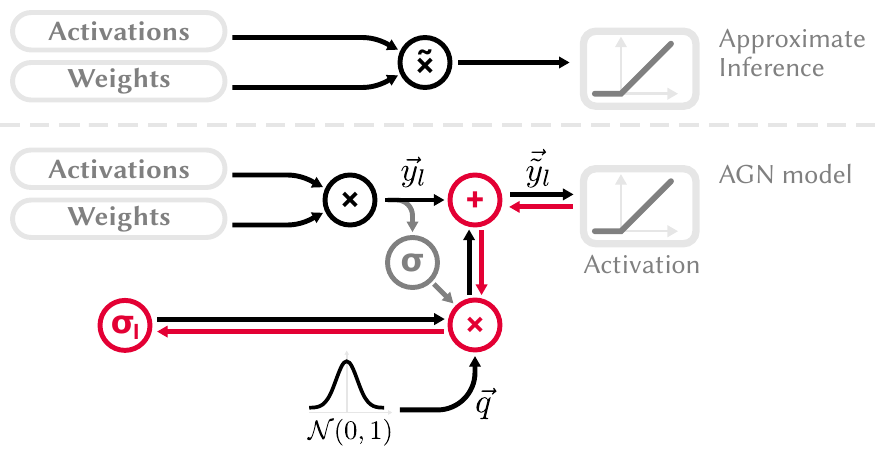}
	\caption{Substituting behavioral simulation of approximate multiplication with parameterized injection of \ac{agn} in a single layer (bias/batch normalization omitted for clarity). Highlighted operations mark backward pass of gradient for $\sigma_l$ w.r.t. task loss. }
	\label{fig:neuron_noise}
\end{figure}
Every layer in our network has a perturbation factor $\sigma_l$ which determines the amount of noise that is added to its accurate output. For each batch, the amount of perturbation is scaled by the standard deviation of the batch as proposed by~\citet{marchisio20}. The relative scaling avoids a situation where the effect of noise of a fixed magnitude would be diminished during training through the network optimizing towards greater pre-activation outputs~\cite{goodfellow16}. To perturb the accurate calculation $\vec{y}_l \in \mathbb{R}^{m}$ of a layer $l$, a tensor with random values of the same dimensions $\vec{q} \sim \mathcal{N}_{m}(0,1)$ is drawn from a normal distribution with a mean of zero and a standard deviation of one. The standard deviation still needs to be adjusted to match the standard deviation of the current batch; to achieve this relative scaling, the tensor is multiplied with the standard deviation of the accurate computation's output for the current batch $\sigma(\vec{y}_l)$ and then weighted with the layer's perturbation factor $\sigma_l$ before being added to the result of the accurate computation as shown in~\Cref{fig:neuron_noise}.
\begin{align}
	\vec{\tilde{y}}_l = \vec{y}_l + \sigma_l \cdot \sigma(\vec{y}_l) \cdot \vec{q}
\end{align}
In the context of this model, the task loss $\mathcal{L}_T$ can be differentiated directly w.r.t $\sigma_l$ using the chain rule:
\begin{align}
\frac{\partial \mathcal{L}_T}{\partial \sigma_l} &= \frac{\partial \mathcal{L}_T}{\partial \vec{\tilde{y}}_l} \cdot \frac{\partial \vec{\tilde{y}}_l}{\partial \sigma_l}\\
 &= \frac{\partial \mathcal{L}_T}{\partial \vec{\tilde{y}}_l} \cdot \sigma(\vec{y}_l) \cdot \vec{q}
\end{align}
Because this formulation is continuous and differentiable, it is possible to optimize all layer perturbations simultaneously---just like all other network parameters---using backpropagation. Computation errors in an early layer can impact the overall accuracy disproportionately as they might alter the results in all subsequent layers; these complex dynamics of propagating errors are already captured by this model because of the chained computation of gradients. The loss function is evaluated based on layers being perturbed by a certain amount of \ac{agn}, resulting in model convergence towards higher robustness to small perturbations. Through the added noise, the output of each individual neuron will be less reliable. Information relevant to the model's task will therefore have to be spread out over more neurons when propagating through the network, similar to the popular Dropout regularization~\cite{srivastava2014}. We assume that this increased robustness will also be beneficial when deploying \acp{am}: because the model has learned to be less reliant on the precise output of individual neurons, it will be able to handle slight deviations in the intermediate results caused by \acp{am} better.\par
Only using the task loss to drive model optimization is not sufficient, however, since $\sigma_l$ could always be driven towards zero. To avoid this, we introduce an additional noise loss $\mathcal{L}_N$ that incentivizes the optimizer to explore solutions with higher perturbation. The noise loss takes into account the current values of $\sigma_l$ as well as the relative cost of each layer $c_l = c(l) / \sum_{l \in L} c(l)$. The relative cost scales the importance of the amount of perturbation in each layer; it is clear that we care most about high values of $\sigma_l$ in layers with a high complexity, while it is preferable to allow for higher relative accuracy in layers that do not contribute much to overall resource consumption. We choose the amount of multiplications in a layer as an easy to implement cost function $c(l)$. The additional noise loss solves the problem of the optimizer converging to solutions with no \ac{agn}, but it creates a new one: The optimizer could now decrease the total loss indefinitely by adding ever-growing amounts of noise to the intermediate results, making the task loss insignificant. We avoid this by upper-bounding the maximum allowable noise loss to some fixed value $\sigma_{\text{max}}$. This gives us the total noise loss as
 \begin{equation}
 	\mathcal{L}_N = - \sum\limits_{l \in L} \min\left\lbrace |\sigma_l|, \sigma_{\text{max}} \right\rbrace \cdot c_l\label{eq:noise_loss}
 \end{equation}
The overall loss $\mathcal{L}$ is then simply the weighted sum of task and noise loss
\begin{equation}
	\mathcal{L} = \mathcal{L}_T + \lambda \cdot \mathcal{L}_N
\end{equation}
where $\lambda$ is a hyperparameter that determines the relative importance of network accuracy and perturbation.~\Cref{eq:noise_loss} can be differentiated w.r.t. $\sigma_l$ as
\begin{equation}
	\frac{\partial \mathcal{L}_N}{\partial \sigma_l} = 
	\begin{cases}
		-c_l, & |\sigma_l| \le \sigma_{\text{max}}\\
		0, &\text{otherwise}
	\end{cases}
\end{equation}
After optimizing the amount of injected noise per layer we arrive at a configuration where each layer's learned robustness to noise $\sigma_l$ has been tuned to maximize the amount of overall \ac{agn} throughout the network while minimizing the degradation of accuracy. Each layer's $\sigma_l$ can be considered a proxy for the actual (but much harder) optimization task: determining how sensitive the network's overall performance is to inaccurate computation results in each layer\par

\subsection{Probabilistic Multi-Distribution Error Model}
\label{ssec:multidistribution}
An important question that arises is how the abstract optimization factor $\sigma_l$ in the \ac{agn} model relates to the error produced by a given \ac{am}. In order to link both, we propose the use of a probabilistic model that treats each error as a random event with certain probabilities; instead of truthfully simulating the error function for each pair of input values, its output can be considered a discrete random variable $Z = e(x,w)$ that maps the outcome of randomly selecting $x$ and $w$ to the error produced by these operands. Determining the standard deviation of $Z$ makes it comparable with the learned perturbation factor $\sigma_l$ that simulates the non-recoverable portion of the approximation error in the \ac{agn} space. Assuming the commonly used 8-bit Integer multipliers, both $x$ and $w$ can take $2^8 = 256$ distinct values for their sample spaces $\Omega_x$ and $\Omega_w$. In total, there are $256^2$ possible combinations of input values in the joint sample space $\Omega_Z = \Omega_x \times \Omega_w$. For any multiplier of interest, we need to know the error (i.e. the difference between accurate and approximate output) for each of these input combinations, from hereon referred to as the multiplier's \emph{error map}. This is the only part of the method that requires at least a high-level model of the hardware to be simulated. If simulating the hardware is expensive, this can be done once and the resulting error map stored for later use.\par
Next, we need to consider that not all $256$ values that $x$ and $w$ might take are equally likely to appear. The relative frequencies of values in the weights tensor and an input sample can be used to model their probability distributions $p_x(x)$ and $p_w(w)$. In practice, this means building a histogram with the count of each possible 8-bit Integer value for both tensors and normalizing it to one. The data-driven construction of operand probabilities means that no assumptions have to be made with regards to their underlying distribution. Given the error for each combination of values and their respective likelihoods, mean and standard deviation of the error can be estimated as
\begin{align}
	\mu_Z &= \sum\limits_{x \in \Omega_x} \sum\limits_{w \in \Omega_w} p_x(x) \cdot p_w(w) \cdot e(x,w)\label{eq:mu}\\
	\sigma_Z^2 &= \sum\limits_{x \in \Omega_x} \sum\limits_{w \in \Omega_w} p_x(x) \cdot p_w(w) \cdot \left(  e(x,w) - \mu_Z \right)^2\label{eq:std}
\end{align}
which describes the mean and standard deviation of the error for a single multiplication. From~\Cref{eq:neuron}, it is clear that the neuron output is the sum over $n$ multiplications. $n$ is the \emph{fan-in} of the neuron, or more simply the number of incoming connections. For \acp{cnn} and most other common \ac{nn} architectures the fan-in is identical for all neurons in a layer. Under the assumption that $x$ and $w$ are independent and identically distributed (an assumption that we will address in more detail in the next paragraph), we can apply the central limit theorem to the growth of the standard deviation. With this, the error mean at the neuron output scales linearly, while the standard deviation scales with the square root of the fan-in:
\begin{align*}
\mu_e &= n \cdot \mu_Z\\
\sigma_e &= \sqrt{n} \cdot \sigma_Z
\end{align*}
The second observation is important because it implies that layers with a higher fan-in will produce an error that is smaller relative to the magnitude of the output value (which grows with $n$), all other factors being equal.\par
So far, we have assumed $x$ and $w$ to be independent and identically distributed. Empirically validating this assumption shows that it holds well for weights, but not for activations. Intuitively, this can be explained by a higher amount of local correlations in the activations. To make this clearer, we can imagine a black and white image passing through a convolutional layer: because pixels of the same color tend to be grouped together, the individual patches are more likely to be all-black or all-white than the global distribution of pixels in the image would suggest. More concisely, the \emph{local} distribution of feature values can deviate strongly from their \emph{global} distribution.\par
\begin{figure}[tb]
	\centering
	\includegraphics[width=\linewidth]{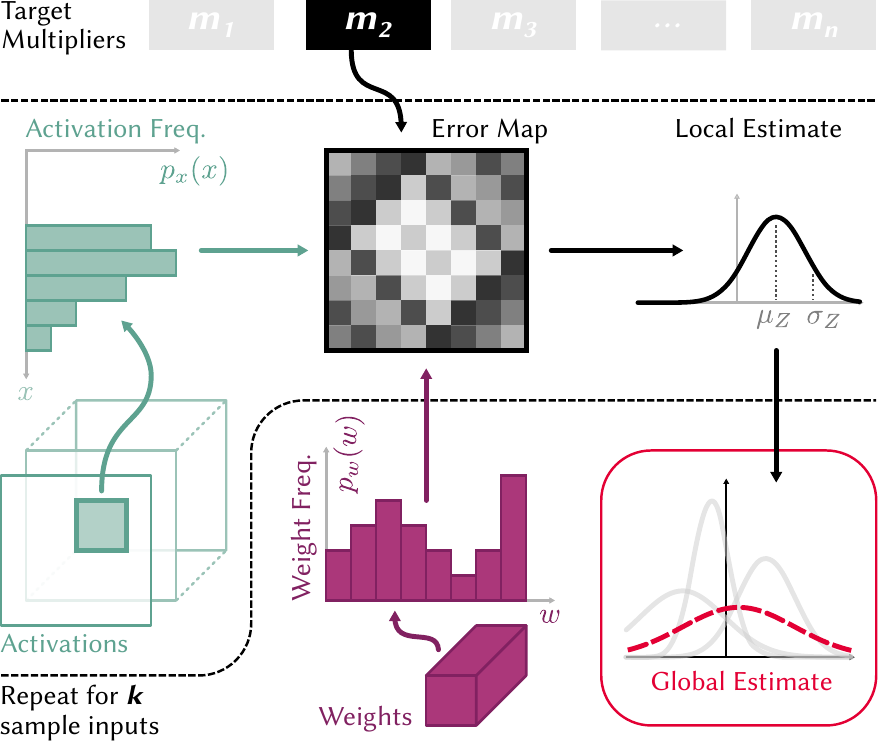}
	\caption{Process of deriving a global estimate of multiplier error properties from multiple local estimates.}
	\label{fig:multidistribution}
\end{figure}
The problem of diverging local and global distributions of operand values can be addressed through sampling; mean and standard deviation are first calculated for several samples of the local distribution. The sample is drawn from the receptive field of a neuron in the target layer, i.e. a sample is either a randomly selected input feature vector for \ac{fc} layers or a patch for convolutional layers. These individual observations are then integrated into an estimate of the global distribution's mean and standard deviation as shown in~\Cref{fig:multidistribution}. We randomly sample $k$ vectors from the layer's input activations. For each input vector, a frequency distribution $p_{x}(x)$ is generated. \Cref{eq:mu,eq:std} are then used to calculate its mean and standard deviation. As an additional benefit, this removes the need to build a global histogram of the input value distribution in favor of building multiple distributions on very small input samples. Calculating the combined mean of the local distributions is simple. Special care needs to be taken when combining group standard deviations: Our method must account for the effect of different means in each sample on the combined standard deviation~\cite{lange18}. 
\begin{align}
	\mu_Z 				&= \frac{1}{k} \sum_{i=1}^{k} \mu_{Z_i}\\
	\sigma_{Z}^2	 	&= \frac{1}{k} \left[\sum_{i=1}^{k}\left(\sigma_{Z_i}^2 + \mu_{Z_i}^2\right) - \frac{1}{k} \left(\sum_{i=1}^{k}\mu_{Z_i}\right)^2\right]
\end{align}
The degree with which this method is applicable is determined by the fan-in $n$ of the respective layer, as convergence of the error towards a normal distribution depends primarily on the fan-in. If layers with a very low fan-in were present, some care would have to be taken to confirm that the \ac{agn} model is still valid.
\subsection{Multiplier Matching}
Our method focuses on Integer \acp{am} with a low bit width which are typically used in low-energy inference settings like edge devices and accelerators. Because of the small amount of operand combinations, a full error map should normally be easy to obtain. To increase the degrees of freedom of the solution, having a large set of different \acp{am} that cover a wide range of accuracy and resource consumption is desirable. As long as these conditions are met, our method is applicable to many different \acp{am} designs.\par
With an optimized configuration of $\sigma_l$ as a measure of a layer's sensitivity, we can match an appropriate \ac{am} to each layer. For every \ac{am} in our search space, we calculate an estimate of the error it produces using the method described in detail in~\Cref{ssec:multidistribution}. The estimate produced by the error model is directly comparable to the learned $\sigma_l$. The multipliers for which the error lies above the accuracy threshold $\sigma_l$ are discarded since they do not produce the required accuracy. From the remaining set of multipliers that produce sufficiently accurate results, we can pick the one that optimizes another metric of interest.
\section{Results and Discussion}
Unless stated otherwise, all experiments were carried out on a host system equipped with an AMD Ryzen 9 3900X CPU and a single nVidia RTX 2080Ti GPU. We implement our search algorithm~\cite{trommer22} as an extension of the popular deep learning framework PyTorch~\cite{pytorch19}. 
\subsection{Multi-Distribution Error Model}
In order to assess the accuracy with which our error model can infer the standard deviation of the approximation error on the layer's output, we evaluate its performance on the layers of a ResNet8 model. The model is trained using the parameters laid out in~\Cref{sec:resnet_cifar}. We do not explicitly evaluate the estimate of the error mean, as it is less relevant to our method for the reasons discussed in~\Cref{sec:noise}. From~\Cref{eq:std}, it should be clear that an accurate estimate of the error mean is a necessary condition for an accurate estimate of the standard deviation. The ground truth for the approximation error is obtained through a behavioral simulation of all 13 unsigned multipliers from the EvoApprox library~\cite{mrazek17}. The simulation generates the approximated pre-activation output of each layer, given its weights and activations. We compare the measured error's standard deviation to the estimate of the standard deviation generated by our probabilistic multi-distribution error model for $k=512$ input samples. The results are also compared to the single-distribution \ac{mc} method discussed in~\cite{marchisio20}. We further include the respective multiplier's \ac{mre} in the evaluation as it is a commonly proposed~\cite{hammad19,delaparra20} single-value metric, used as an indication of multiplier performance. The results of our evaluation can be found in~\Cref{tab:errormodels}.\par
\begin{table}[tb]
	\caption{Comparison of predictive methods for multiplier error standard deviation on ResNet8 layers}
	\footnotesize
	\begin{tabularx}{\linewidth}{lYY}
		\toprule
		Error Model 								& Pearson Correlation 	& Median Relative Error $\pm$ Interquartile Range\\
		\midrule
		Multiplier MRE~\cite{hammad19}				& 0.546			& n.a. \\
		Single-Distribution MC~\cite{marchisio20}	& 0.767			& $\left(42.9 \pm 53.2\right)\%$ \\
		Probabilistic Multi-Dist. (ours)			& 0.997			& $\left(4.6 \pm 8.8\right)\%$ \\
		\bottomrule
	\end{tabularx}
	\label{tab:errormodels}
\end{table}
We find that the results of the Single-Distribution \ac{mc} method are very similar to the results produced by using our probabilistic method \emph{without} accounting for differing local distributions, i.e. based on the global frequency distribution of activation values alone. Single-Distribution \ac{mc} and our probabilistic model converging to similar results in this scenario is very plausible, given that the former method is a \ac{mc} simulation of exactly the same process that is analytically described in the latter. This suggests that taking local divergence of the input activations into account provides a significant boost in the predictive performance of our model, compared to methods that assume the global distribution to be present for all operations. The observed values for the standard deviation occupy a very wide numerical range of approximately 5 orders of magnitude. On such a large range, a median relative error of \SI{4.6}{\percent} is negligible, which is reflected in the almost perfect correlation of $0.997$.\par
Our evaluation shows that the predictive performance of the \ac{mre} for the error's standard deviation is very limited. This can be explained by the fact that the \ac{mre} is only a metric of a multiplier performance over the entire numerical range and for operands with equal probabilities. It does not distinguish between systematic and non-systematic portions of the error and can include neither knowledge about the different likelihoods of operands in \ac{nn} models nor the impact of a layer's fan-in on the aggregate error at the layer output.
\subsection{ResNet on CIFAR-10}
\label{sec:resnet_cifar}
\begin{table}[tb]
	\caption{Comparison of energy reduction and accuracy loss for different methods}
	\footnotesize
	\begin{tabularx}{\linewidth}{llYY}
		\toprule
		Model & Method & Energy Reduction & Top-1 Accuracy Loss [p.p.]\\
		\midrule
		ResNet8		& ALWANN~\cite{mrazek19}				& \SI{30}{\percent}	& 1.7 \\
		{}			& Uniform Retraining~\cite{delaparra20}	& \SI{58}{\percent}	& 0.9 \\
		{}			& Gradient Search (ours)				& \SI{70}{\percent}	& 0.5 \\
		\midrule
		ResNet14	& ALWANN~\cite{mrazek19}				& \SI{30}{\percent}	& 0.9 \\
		{}			& Uniform Retraining~\cite{delaparra20}	& \SI{57}{\percent}	& 0.9 \\
		{}			& Gradient Search (ours)				& \SI{75}{\percent}	& 0.9 \\
		\midrule
		ResNet20	& LVRM~\cite{tasoulas20}				& \SI{17}{\percent}	& 1.0 \\
		{}			& Uniform Retraining~\cite{delaparra20}	& \SI{53}{\percent}	& 0.7 \\
		{}			& Gradient Search (ours)				& \SI{71}{\percent}	& 0.9 \\
		\midrule
		ResNet32	& Gradient Search (ours)				& \SI{79}{\percent}	& 0.8 \\
		\bottomrule
	\end{tabularx}
	\label{tab:resnet}
\end{table}
In line with previous works~\cite{mrazek19, delaparra20}, our method is evaluated on several sizes of the ResNet~\cite{he16} architecture, trained on the CIFAR-10 data set~\cite{krizhevsky09} with 36 8-bit unsigned multipliers from the EvoApprox library~\cite{mrazek17} as the \ac{am} search space. Relative power numbers were generated using the \texttt{pdk45\_pwr} property of the EvoApprox multipliers, normalized to the power consumption of the reference accurate multiplier and weighted with the number of multiplications in each layer.\par
The training and augmentation procedure laid out in the original work is used to generate a floating-point reference model. Next, \ac{qat} is carried out to obtain a baseline model that is quantized to 8 Bit. On the quantized baseline, we uniformly initialize each layer's perturbation factor $\sigma_l$ to $0.1$ with a cap at $\sigma_{\text{max}}=0.5$. Layer perturbation factors and other network parameters are then jointly optimized for 30 epochs using the \ac{sgd} optimizer with an initial learning rate of $1\times 10^{-2}$ and a decay of $0.9$ after every 10 epochs. After matching \acp{am} with appropriate accuracy to each layer, we retrain the entire network using behavioral simulation of the selected \acp{am} for another 5 epochs with an initial learning rate of $1\times 10^{-3}$ and a decay of $0.9$ after every other epoch. During this phase, the \ac{ste}~\cite{bengio13, he2018} is used to derive valid gradients for the \acp{am}. We repeat the Gradient Search and retraining several times while varying the $\lambda$ parameter between $0$ and $0.6$ in steps of $0.05$. By adjusting the value of $\lambda$, we can generate several solutions, each of which strikes a different balance between accuracy and perturbation (and thus, by extension, energy consumption).\par
\begin{figure}[tb]
	\centering
	\includegraphics[width=\linewidth]{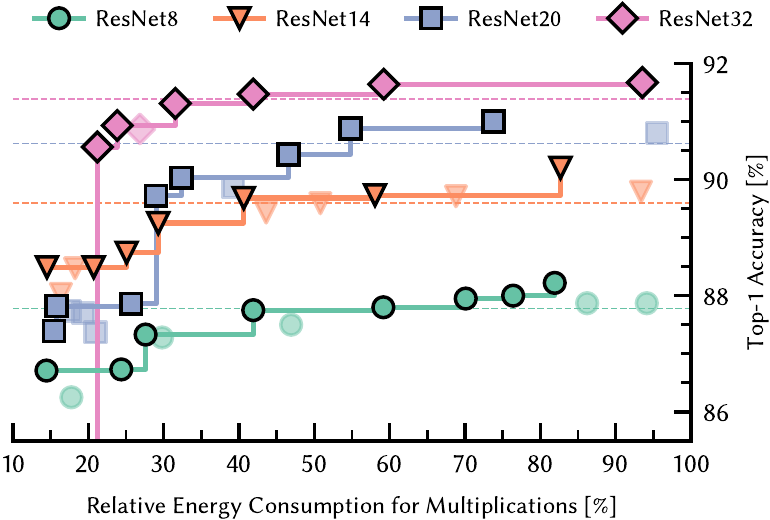}
	\caption{Pareto Front and dominated points of energy and accuracy for different sizes of ResNet~\cite{he16} on the CIFAR-10 data set~\cite{krizhevsky09}. Horizontal line marks 8-bit quantized baseline accuracy.}
	\label{fig:plot_pareto}
\end{figure}
In the comparison of results in~\Cref{fig:plot_pareto} it can be seen that the accuracy is above the baseline for all ResNet variants for an energy reduction of up to 45\%. We attribute this to the perturbation acting as a form of learnable regularization; since all parameters are trained in the Gradient Search phase, the network converges towards a configuration that is both more resilient to approximation and generalizes better to unseen data. The drop in accuracy for higher degrees of approximation becomes increasingly steep for deeper models, most likely due to the accumulating effect of the approximation error.\par
\begin{figure}[tb]
	\centering
	\includegraphics[width=\linewidth]{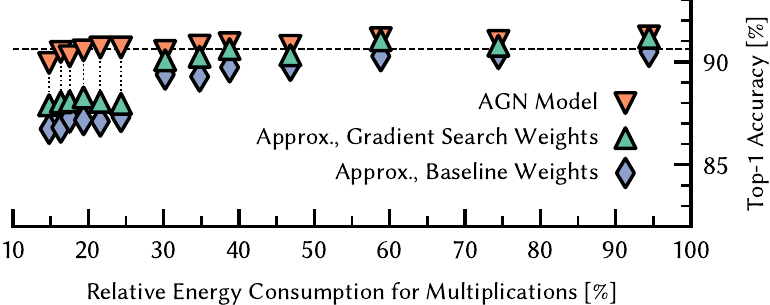}
	\caption{Comparison of perturbed accuracy and accuracy after retraining with weights and biases of AGN model and baseline  model for ResNet20. Horizontal line marks 8-bit quantized baseline accuracy.}
	\label{fig:plot_retrain_error}
\end{figure}
The comparison of the accuracy in the \ac{agn} space (\enquote{AGN Model}) with the accuracy that is achieved after retraining using behavioral simulation (\enquote{Approx., Gradient Search Weights}) in~\Cref{fig:plot_retrain_error} reveals another interesting property: the accuracy of a retrained model under approximation is very similar to the accuracy exhibited by the model that is perturbed using \ac{agn} up to energy savings of around \SI{60}{\percent}. For more aggressive approximation, the relationship between both values deteriorates up to a gap of several percentage points. These models suffer from a significant degradation of accuracy under approximation, while the same is not true for the same model perturbed with comparable amounts of \ac{agn}. This can be interpreted as a shortcoming of the \ac{agn} model: \acp{am} with very low energy consumption are more likely to produce localized error patterns which are insufficiently captured by the \ac{agn} model. The \ac{agn} model assumes an even spread of errors over all neurons. It is likely that a structured error has a more adverse effect on the propagation of information for some specific connections, widening the gap between predicted and achieved accuracy.\par
\Cref{fig:plot_retrain_error} also provides some insight into the question of whether training in the \ac{agn} space has a positive carryover to model accuracy when deploying \acp{am}. To make the impact of training with \ac{agn} on the achieved accuracy visible, we repeat the retraining using behavioral simulation based on the weights and biases of the baseline model (\enquote{Approx., Baseline Weights}) instead of the weights and biases learned during the Gradient Search phase. We can see that the models trained using \ac{agn} consistently achieve higher accuracies after retraining using behavioral simulation compared to models that directly attempt approximate retraining on the baseline model. This indicates that models trained using \ac{agn} have converged to a configuration that is more robust to the errors produced by approximation.\par
A comparison of the loss of accuracy and reduction in energy consumption to other state-of-the-art methods can be found in~\Cref{tab:resnet}. In addition to improved energy efficiency, our algorithm adds little runtime overhead. The Gradient Search only takes between \num{6} \si{\minute} for ResNet8 and \num{16} \si{\minute} for ResNet32. Across all networks, this puts the overhead between \SIrange{41}{45}{\percent} of the time taken to train the respective floating-point reference networks---much lower than the hours to days required by~\citet{mrazek19}. The multiplier matching algorithm that generates the estimated error standard deviation for each combination of layer and multiplier completes in around one minute for all surveyed networks on our target system.
\subsection{VGG16 on Tiny ImageNet}
\begin{figure}[tb]
	\centering
	\includegraphics[width=\linewidth]{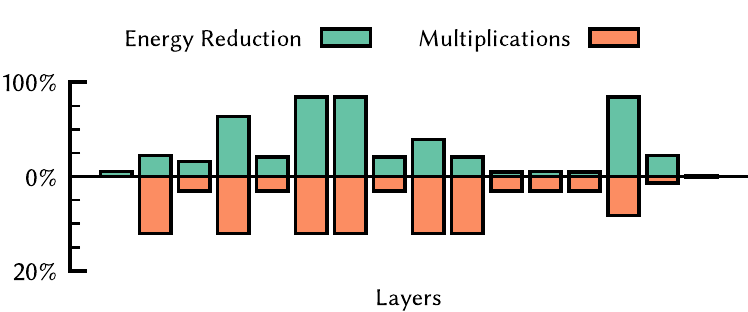}
	\caption{Comparison of energy reduction per layer and relative multiplications for VGG16, based on output configuration of unsigned EvoApprox multipliers}
	\label{fig:vgg16_config}
\end{figure}
\begin{table}[tb]
	\caption{Comparison of homogeneous and heterogeneous solutions for VGG16~\cite{simonyan15} architecture on the Tiny ImageNet dataset~\cite{le2015}}
	\footnotesize
	\begin{tabularx}{\linewidth}{lYY}
		\toprule
		Configuration 								& Energy Reduction  	& Top-5 Val. Accuracy\\
		\midrule
		Baseline									& n.a.					& \SI{80.6}{\percent} \\
		AGN Model, $\lambda = 0.3$					& n.a.					& \SI{80.2}{\percent} \\
		\midrule
		Uniform Retraining, \texttt{mul8s\_1KVB}	& \SI{3.5}{\percent}	& \SI{80.3}{\percent} \\
		Heterogeneous, signed						& \SI{11.6}{\percent}	& \SI{80.5}{\percent} \\
		\midrule
		Uniform Retraining, \texttt{mul8u\_19DB}	& \SI{52.7}{\percent}	& \SI{79.6}{\percent} \\
		Uniform Retraining, \texttt{mul8u\_185Q}	& \SI{52.7}{\percent}	& \SI{79.7}{\percent} \\
		Heterogeneous, unsigned						& \SI{52.6}{\percent}	& \SI{80.1}{\percent} \\
		\bottomrule
	\end{tabularx}
	\label{tab:vgg_tinyimagenet}
\end{table}
To demonstrate the scalability of our method, we apply it to the more complex Tiny ImageNet dataset~\cite{le2015}. Tiny ImageNet is a smaller version of the ImageNet challenge~\cite{deng09} in which the number of classes is reduced from 1000 to 200. Each class contains 500 training and 50 validation images and images are down-scaled from $224 \times 224$ pixels to $64 \times 64$ pixels. The VGG16 \ac{cnn} architecture~\cite{simonyan15} with additional batch normalization is used as a reference architecture. The \ac{sgd} optimizer is used for all training runs as we find that the popular ADAM optimizer~\cite{kingma14} does not produce satisfactory results when optimizing the layer perturbations. Only a single run of Gradient Search with $\lambda=0.3$ and an initial perturbation factor of $\sigma_l = 0.025$ for all layers is performed to account for the larger dataset. We also lower the amount of epochs for the Gradient Search and approximate retraining to 9 and 2 respectively. Based on the result in the \ac{agn} space, both the signed and unsigned 8-bit multipliers from the EvoApprox library are used separately as search spaces for matching~\ac{am} instances.\par
Results in~\Cref{tab:vgg_tinyimagenet} show that heterogeneous configurations outperform uniform solutions when trading off energy savings and loss of accuracy. Given the more complex classification task, we do not find an improvement over the accuracy of the baseline model anymore, both in the \ac{agn} space as well as after deployment of \acp{am} and retraining. The much lower reduction of energy consumption when using signed multipliers can be attributed their lower overall energy reduction for similar performance as well as the smaller search space of only 13 available signed 8-bit multipliers.\par
For the unsigned solution, the search procedure identifies 13 different \acp{am} from the search space. The comparison of layer complexity and the respective reduction in energy consumption per layer in~\Cref{fig:vgg16_config} shows that the Gradient Search and subsequent multiplier assignment deploys the highest degrees of approximation on the network's inner layers with high amounts of multiplications. In contrast, particularly the first and last layers are assigned highly accurate hardware instances. A relatively high sensitivity of these layers is in line with common heuristics in non-uniform quantization schemes~\cite{wang19}.\par
We confirm that this effect is consistent by repeating the experiment with other \ac{cnn} models. Performing the same optimization procedure on AlexNet~\cite{krizhevsky12} and MobileNetV2~\cite{sandler18} models yields similar results: Inner layers with higher computational complexity are assigned less accurate hardware instances with up to \SI{84.4}{\percent} reduction in energy consumption. In contrast, highly accurate \acp{am} with a reduction between \SI{1.3}{\percent} and \SI{5.4}{\percent} are mapped to the first and last layers of each model.
\section{Conclusion}
In this work, we have demonstrated the feasibility of combining heterogeneous approximation and retraining for \acp{nn}. The use of \ac{agn} as a surrogate model for behavioral simulation is motivated from the mathematical formulation of \acp{nn}. Our findings suggest that there is a strong connection between perturbation of neural networks with \ac{agn} and low to medium degrees of approximation. This connection, however, weakens for higher degrees of approximation. Whether the \ac{agn} model can be adapted to capture more structured errors without sacrificing generalizability remains to be answered by further research. Despite this, we show that the trade-off between energy and accuracy can be improved consistently and significantly by considering the varying robustness of different parts of a \ac{nn} during the deployment of \acp{am}. As a means to tackle the enormous search space of heterogeneous \ac{am} assignment, the flexibility of the \ac{agn} model is used to abstract over many different \acp{am} with only a single parameter. This parameter is differentiable, so it can be learned using backpropagation. Learning the different degrees of robustness to errors during network training is the key to efficiently exploring different solutions, especially for deeper networks. To connect \ac{agn} to concrete hardware, we introduced a probabilistic model of the approximation error. This model allows for an accurate prediction of the noise properties which the \ac{am} error will exhibit on any given layer without having to rely on time-consuming simulations. Through combining these methods, we provide a path towards \ac{nn} hardware that makes inference both accurate and power-efficient.

\newpage
\bibliographystyle{ACM-Reference-Format}
\bibliography{bibliography}

\end{document}